# Long Text Generation Challenge

**Nikolay Mikhaylovskiy**
Higher IT School, Tomsk State University, Tomsk, Russia, 634050
NTR Labs, Moscow, Russia, 129594
nickm@ntr.ai

## Abstract

We propose a shared task of human-like long text generation, *LTG Challenge,* that asks models to output a consistent human-like long text (a Harry Potter generic audience fanfic in English), given a prompt of about 1000 tokens. We suggest a novel statistical metric of the text structuredness, GloVe Autocorrelations Power/ Exponential Law Mean Absolute Percentage Error Ratio (GAPELMAPER) and a human evaluation protocol. We hope that *LTG* can open new avenues for researchers to investigate sampling approaches, prompting strategies, autoregressive and non-autoregressive text generation architectures and break the barrier to generate consistent long (40K+ token) texts.

## 1 Task Overview

The human-like long text generation (*LTG*) task asks models to output a consistent human-like long text (a Harry Potter generic audience fanfic in English), given a prompt of about 1000 tokens. The text will be evaluated by an automated metric GloVe Autocorrelations Power/Exponential Law Mean Absolute Percentage Error Ratio (GAPELMAPER), described in Section 3.1, and a human evaluation protocol described in Section 3.2.

## 2 Motivation

Autoregressive probabilistic large language models (LLMs) have been the cornerstone for solving every task in computational linguistics through few-shot learning (Brown et al., 2020) or prompt engineering (Sahn et al., 2021). Many users now interact with such models as ChatGPT, Claude, or Google Bard in chat setting regularly. However, these models still have many deficiencies. Despite the targeted effort, they can generate false information, propagate social stereotypes, and produce toxic language (Taori et al., 2023).

The LLM deficiency we particularly want to attack is their inability to produce a human-grade long text. Current autoregressive language models fail to catch long-range dependencies in the text consistently. Large language models such as GPT-3 (Brown et al., 2020), LLaMA (Touvron et al.), ALPACA (Taori et al.) push the boundary of "short text" rather far, but do not remove the problem. Commercial instruction-following language models such as ChatGPT, GPT-4, Claude and Google Bard are targeted at the use in a dialogue (and probably that is not for nothing). They generate a limited number of tokens per user input, and only generate further text after additional prompting. While the autoregressive window for commercial models reaches 32K tokens, which is a lot, it is not clear if that allows them to generate long coherent texts.

Modeling long texts requires many distinct abilities compared to short texts (Guan et al., 2022), including (1) commonsense reasoning regarding characters' reaction and intention, and knowledge about physical objects (e.g., "river") and abstract concepts (e.g., "irony"); (2) modeling discourse-level features such as inter-sentence relations (e.g., causality) and global discourse structures (e.g., the order of events); and (3) the generation coherence and controllability, which require both maintaining a coherent plot and adhering to controllable attributes (e.g., topics).

Mikhaylovskiy and Churilov (2023) have recently studied autocorrelations in long texts using pretrained word vectors. That allowed to study a wide range of autocorrelation distances in human-written and model-generated texts and show that the autocorrelations in human-written literary texts decay according to power laws on distances from 10 to 10000 words independently from the language. On the other hand, the behavior of autocorrelations decay in generated texts is quantitatively and often qualitatively different from the literary texts. Large language models often exhibit Markovian (Markov, 1913) behavior with exponential autocorrelations decay.

Several authors have shown theoretically and empirically (Lin and Tegmark, 2017, Alvarez-Lacalle et al., 2006) that the power law autocorrelations decay is closely connected with the hierarchical structure of texts. Indeed, the hierarchical structure of, for example, Leo Tolstoy's War and Pease consists of at least 7 levels: the whole novel, books, parts, chapters, paragraphs, words, and letters. There are strong reasons to think that this structure reflects an important aspect of human thinking: people do not generate texts autoregressively. Writing a long text requires some thinking ahead, and going back to edit previous parts for consistency. This going back and forth can be reflected by navigating a tree-like structure. The autoregressive nature of the current state-of-the-art models does not reflect this, for example, S4 model (Gu et al., 2021) exhibits clear exponential autocorrelations decay (Mikhaylovskiy and Churilov, 2023).

We hope that this challenge can gain interest from the NLG community and advance sampling approaches, prompting strategies, autoregressive and non-autoregressive text generation architectures and other subfields of text generation.

## 3 Task Description

Formally, the task of ConvSumX Challenge asks participants to provide a system that can output a consistent human-like long text (a Harry Potter generic audience fanfic), given a prompt of about 1000 tokens.

We employ both automatic and human evaluation, described below, to evaluate the quality of the texts.

### 3.1 GloVe Autocorrelations Power/ Exponential Law Mean Absolute Percentage Error Ratio (GAPEL-MAPER) Metric

Suppose we have a sequence of $N$ vectors $V_i \in R^d, i \in [1, N]$. Autocorrelation function $C(\tau)$ is the average similarity between the vectors as a function of the lag $\tau = i - j$ between them. The simplest metric of vector similarity is the cosine distance $d(V_i, V_j) = \cos\angle(V_i, V_j) = \frac{V_i \cdot V_j}{\|V_i\|\|V_j\|}$, where · is a dot product between two vectors and $\|\ \|$ is an Euclidean norm of a vector. Thus,

$$C(\tau) = \frac{1}{N-\tau} \sum_{i=1}^{N-\tau} \frac{V_i \cdot V_{i+\tau}}{\|V_i\|\|V_{i+\tau}\|}. \quad (5)$$

|  | Power law MAPE | Exp law MAPE | GAPEL-MAPER |
|---|---|---|---|
| The Adventures of Tom Sawyer | 0.21 | 0.55 | 0.38 |
| The Republic | 0.13 | 0.38 | 0.34 |
| Don Quixote | 0.20 | 0.44 | 0.45 |
| War and Peace | 0.09 | 0.42 | 0.21 |
| Critique of Pure Reason | 0.14 | 0.25 | 0.56 |
| The Iliad | 0.19 | 0.54 | 0.35 |
| Moby-Dick or, The Whale | 0.15 | 0.47 | 0.32 |
| S4 generated text | 0.062 | 0.050 | 1.24 |

Table 1: MAPE of power and exp law approximations of texts in English, and resulting GAPELMAPER

$C(\tau)$ ranges from $-1$ for perfectly anticorrelated sequence (for $\tau = 1$ and $d = 1$ that would be $1, -1, 1, -1$ etc.) to 1 for a perfectly correlated one (for $\tau = 1$ and $d = 1$ that would be $1, 1, 1, 1$ etc.).

A distributional semantic assigns a vector to each word or context in a text. Thus, a text is transformed into a sequence of vectors, and we can calculate an autocorrelation function for the text. Two distributional semantics approaches have been proposed for word-level autocorrelation computations: Alvarez-Lacalle et al. (2006) proposed a bag-of-words (BOW) model, and Mikhaylovskiy and Churilov (2023) have suggested the use of pretrained GloVe (Pennington et al., 2014) vectors. Unlike BOW, which only allows measuring long distance correlations, the latter approach allows to measure autocorrelations at any word distance starting with 1. Thus, we suggest using GloVe for autocorrelation measurement.

Mikhaylovskiy and Churilov (2023) have found that autocorrelations in long human-written texts decay according to a power law at ranges from 10 to 10000 words. We suggest measuring the structuredness of a generated text by comparing how well the autocorrelations decay is approximated by power law and exponential law. To do so, one can compute autocorrelations in this range, approximate these points by a straight line in log-log and log-linear coordinates using the least squares regression and evaluate the goodness of fit of these regressions by MAPE (Mean Absolute Percentage Error). The ratio of these two errors constitute a metric we call GloVe Autocorrelations Power/Exponential Law Mean Absolute Percentage Error Ratio (GAPELMAPER):

$$\text{GAPELMAPER} = \frac{MAPE_{power}}{MAPE_{exp}}$$

GAPELMAPER less than 1 means that the autocorrelations decay according to a power law and the text is structured in a way. GAPELMAPER more than 1 means that the autocorrelations decay according to a exponential law and the text is unstructured. As a matter of example, we take Table 3 from Mikhaylovskiy and Churilov (2023) and compute GAPELMAPER in Table 1.

### 3.2 Human Evaluation Approach

A single number is not enough to evaluate the quality of a long text. We adopt multiple human evaluation metrics to better measure model performance. Similarly to Kryscinski et al. (2019), we ask annotators to rate the texts across four dimensions: relevance (of topics in the text to the expected ones), consistency (alignment between the parts of the text), fluency (quality of individual sentences), and coherence (quality of sequence of sentences). Each text will be rated by 5 distinct judges with the final score obtained by averaging the individual scores. To simplify evaluation, each judge may be provided by a human-written text of a similar length starting with the same tokens (fanfic automatically translated into English).

### 3.3 Protocol

We propose the following schedule:
- **Phase 1** (from Sep, 2023): The shared task is announced at the INLG 2023 conference, and the data are available on the shared task website; participants can register to the task.
- **Phase 2** (from Dec, 2023): The leaderboard is open; participants can submit their systems to the organizers and the online leaderboard keeps updating the best performance using automatic evaluation metrics.
- **Phase 3** (from Mar, 2024): The submission is closed; organizers conduct manual evaluation.
- **Phase 4** (Jul, 2024): The LTG Challenge shared task is fully completed. Organizers submit participant reports and challenge reports to INLG 2024 and present at the conference.

For fairness and reproducibility, participants should specify what and how external resources are used in their system reports. In Phase 3, after the submission deadline, the organizers will start to evaluate summaries generated by final submitted models with the help from linguistic experts.

Please note that the above schedule can be modified accordingly when the schedule of INLG 2024 is released. The leaderboard and the detailed schedule will be announced on the shared task website.

## 4 Related work

Shaham et al. (2022) introduced SCROLLS, a suite of tasks that require reasoning over long texts. It includes earlier introduced works of Huang et al. (2021), Chen et al. (2022), Zhong et al. (2021), Dasigi et al. (2021), Kočiský et al. (2018), Pang et al. (2022), and Koreeda and Manning (2021). While all are related to long texts, none of these datasets and tasks asks to generate a long text.

Gehrmann et al. (2021) introduced GEM, a living benchmark for natural language Generation (NLG), its Evaluation, and Metrics. GEM provides an environment in which models can easily be applied to a wide set of tasks and in which evaluation strategies can be tested and consists of 11 datasets/tasks. Tay at al. (2020) proposed Long Range Arena, a suite of tasks consisting of sequences ranging from 1K to 16K tokens, encompassing a wide range of data types and modalities such as text, natural, synthetic images, and mathematical expressions requiring similarity, structural, and visual-spatial reasoning. None of these tasks asks to generate a long text as well.

Very recently Köksal et al. (2023) introduced the LongForm dataset, which is created by leveraging English corpus examples with augmented instructions. No evaluation protocol or competition is suggested in the cited paper.

The most similar effort to ours was most likely made by Guan et al. (2022), who proposed a story-centric benchmark named LOT for evaluating Chinese long text modeling. The benchmark aggregates two understanding tasks and two generation tasks. The authors constructed new datasets for these tasks based on human-written Chinese stories. Unlike our proposal, LOT benchmark is limited to texts hundreds of words long, and Chinese language.

## 5 Conclusion

We propose the LTG Challenge to address the task of long text generation, with the hope that it can open new avenues for researchers to investigate sampling approaches, prompting strategies, autoregressive and non-autoregressive text generation architectures and break the barrier to generate consistent long (40K+ token) texts, and the frontier of text generation can be pushed further.